\documentclass[conference]{IEEEtran}
\IEEEoverridecommandlockouts
\usepackage{cite}
\usepackage{amsmath,amssymb,amsfonts}
\usepackage{algorithmic}
\usepackage{graphicx}
\usepackage{textcomp}
\usepackage[T1]{fontenc}    
\usepackage{nicefrac}       
\usepackage{multirow}
\usepackage{verbatim}
\usepackage{mathtools}
\usepackage{hyperref}
\usepackage[normalem]{ulem}
\DeclarePairedDelimiter{\nint}\lfloor\rceil

\usepackage{microtype}
\usepackage{subfigure}
\usepackage{booktabs} 
\usepackage{pifont}

\usepackage[dvipsnames]{xcolor}

 \usepackage{textcomp}

\def\BibTeX{{\rm B\kern-.05em{\sc i\kern-.025em b}\kern-.08em
    T\kern-.1667em\lower.7ex\hbox{E}\kern-.125emX}}

\newcommand{\floatfp}{{\sc float32}}
\newcommand{\bfloat}{{\sc bfloat16}}
\begin{document}

\title{Photonic Accelerators for Image Segmentation in\\Autonomous Driving and Defect Detection}

\author{
\IEEEauthorblockN{Lakshmi Nair*, David Widemann, Brad Turcott,\\ Nick Moore, Alexandra Wleklinski, Darius Bunandar}\thanks{*Email: lakshmi@lightmatter.co}
\IEEEauthorblockA{\textit{Lightmatter Inc.} \\
Boston, Massachusetts, USA 
}
\and
\IEEEauthorblockN{Ioannis Papavasileiou$^{\dagger}$, Shihu Wang, Eric Logan}\thanks{$\dagger$Email:Papavasii@corning.com}
\IEEEauthorblockA{\textit{Corning Inc.} \\
Corning, New York, USA 
}
}


\maketitle

\begin{abstract}
Photonic computing promises faster and more energy-efficient deep neural network (DNN) inference than traditional digital hardware. Advances in photonic computing can have profound impacts on applications such as autonomous driving and defect detection that depend on fast, accurate and energy efficient execution of image segmentation models. In this paper, we investigate image segmentation on photonic accelerators to explore: a) the types of image segmentation DNN architectures that are best suited for photonic accelerators, and b) the throughput and energy efficiency of executing the different image segmentation models on photonic accelerators, along with the trade-offs involved therein. Specifically, we demonstrate that certain segmentation models exhibit negligible loss in accuracy (compared to digital \floatfp{} models) when executed on photonic accelerators, and explore the empirical reasoning for their robustness. We also discuss techniques for recovering accuracy in the case of models that do not perform well. Further, we compare throughput (\textit{inferences-per-second}) and energy consumption estimates for different image segmentation workloads on photonic accelerators. We discuss the challenges and potential optimizations that can help improve the application of photonic accelerators to such computer vision tasks.
\end{abstract}


\section{Introduction}

Semantic segmentation of imagery is an important computer vision task for various applications ranging from robotics \cite{hurtado2022semantic} and autonomous driving \cite{cakir2022semantic} to defect detection during manufacturing processes \cite{roberts2019deep}, with an increasing number of models being released for such tasks even today \cite{kirillov2023segment}. Over the past decade, DNNs for image segmentation have evolved from convolutional neural networks (CNNs) \cite{he2015deep, https://doi.org/10.48550/arxiv.1409.1556, zhou2018semantic} to transformer-based networks that outperform CNNs on these tasks \cite{liu_swin_2021}. However, the sizes of these DNNs and the amount of computation required for training such models have outpaced Moore's law \cite{openAi}. In order to address the problems of growing model latency and energy usage, researchers have turned to the use of photonic computing for quickly and efficiently performing matrix-vector products (MVPs), which are the primary computations in DNNs \cite{nahmias2019photonic, peserico2023integrated}. While existing work has explored the use of FPGA (field-programmable gate array) based accelerators for autonomous driving, for achieving low latency \cite{ghielmetti2022real, kim2020accelerator, de2006fpga}, recent advances in photonic computing are yet to be explored in this domain. 

\begin{figure}[t]
\centering
\includegraphics[width=0.48\textwidth]{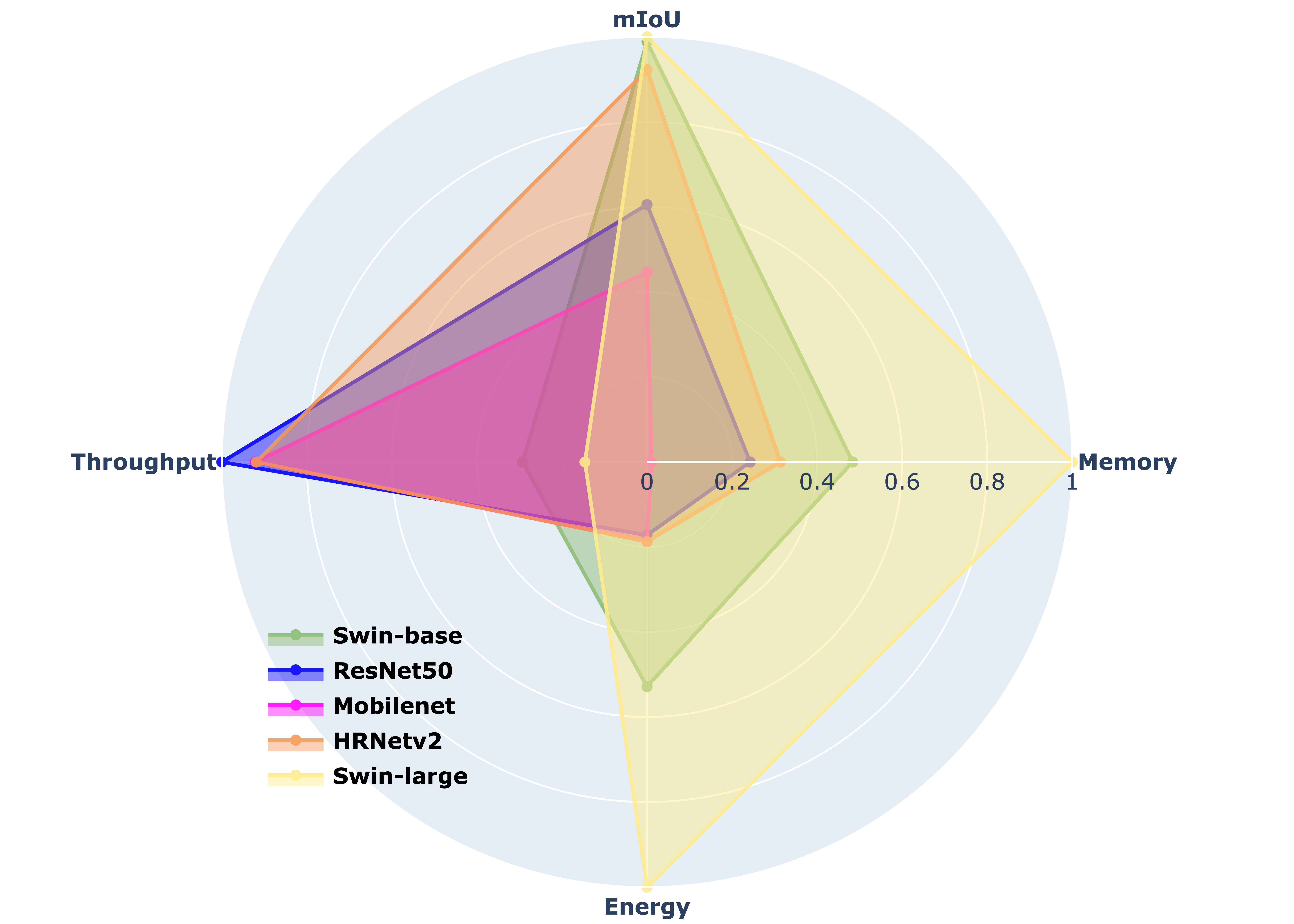}
\caption{The relative mIoU (accuracy), memory, energy cost, and throughput of five DNN models performing image segmentation on three datasets using a photo-core (photonic compute unit). Vision transformers, like Swin (maskformer), are more accurate out-of-the-box at the cost of requiring more energy and memory than their CNN counterparts.}
\label{fig:radar}
\end{figure}

In this paper, we focus on image segmentation workloads on photonic accelerators for autonomous driving and defect detection in manufacturing -- two important applications that require fast, accurate and real-time image segmentation capabilities \cite{holder2022efficient, stavropoulos2020vision}. For this work, we do not contribute our own accelerator design. Instead, we focus on the implications of executing image segmentation workloads on existing photonic devices, by basing our work on a recently proposed photonic  compute core, called the \textit{photo-core} \cite{demirkiran2022electrophotonic}. While there is some variability among photonic accelerator designs, we believe that the findings of this work will offer common insights and challenges associated with executing image segmentation DNNs on photonic accelerators. Specifically, we make the following contributions: a) we analyze recent image segmentation DNNs on the photo-core, alongside techniques such as fine-tuning for accuracy recovery; b) we identify the DNN architectures that achieve good out-of-the-box (OOB) accuracy compared to \floatfp{}. We further investigate the empirical reasons for the accuracy robustness of some architectures over others, in order to guide the design of future DNNs specialized for photonic accelerators; c) we investigate the relative throughput and power estimates for the image segmentation models on the photo-core to characterize the relative trade-offs for the DNNs on photonic devices. To our knowledge, this is the first work to investigate image segmentation on photonic accelerators, with the goal of providing insights for computer vision on photonic hardware.


\begin{figure*}[t]
\centering
\includegraphics[width=0.9\textwidth]{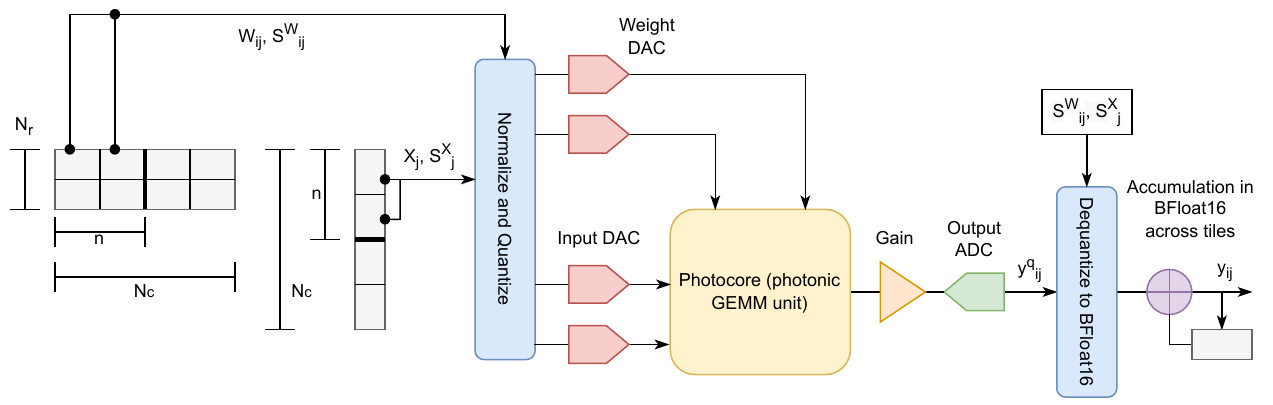}
\caption{High-level overview of applying ABFP with the photo-core unit. Input weights and activations are scaled at per-vector granularity and the quantized matrices are mutliplied on the photocore. The output result is then amplified, quantized at the output ADC, and de-quantized for accumulation in digital.}
\label{fig:abfp_flow}
\end{figure*}

\section{Related Work}

\subsection{ML applications on accelerators}
Recently, digital machine learning accelerators have attained significant interest owing to the potential speed benefits that they offer for ML workloads. Since the primary computational component of DNNs is matrix multiplication, specialty ASICS such as Nvidia's Tensor Cores \cite{nvidia2017v100} and Google's TPU \cite{jouppi2017datacenter} have been developed to accelerate matrix multiplication. However, the power consumption of these systems has been increasing with each generation, and their throughput has not improved correspondingly \cite{6307773}. As a result, recent approaches have proposed photonic accelerator designs that demonstrate higher throughput with lower energy profiles, offsetting the disadvantages of more traditional systems \cite{demirkiran2022electrophotonic,liu2019holylight,mehrabian2018pcnna}. However, photonic designs present numerous challenges associated with noisy electro-optical components (e.g., thermal, shot noise) alongside their reduced precision that impact the performance of DNNs \cite{paolini2022photonic}. 

Prior work has looked into performing high-speed image classification \cite{ashtiani2022chip, wang2022integrated} and natural language processing (NLP) \cite{valensise2022large} on photonic accelerators. For image classification, prior work demonstrated inference speeds of under 570 ps with high accuracy, for classification of hand-written letters \cite{ashtiani2022chip}. Their work used a clock-less processing of data thus eliminating analog-to-digital conversions to improve throughput and energy efficiency. Further work demonstrated an integrated metasystem, in contrast to conventional photonics integrated circuits, for high accuracy and throughput on the MNIST dataset \cite{wang2022integrated}. For NLP, prior work demonstrated a novel design of photonic extreme learning machines (PELM) for performing sentiment analysis on the Image Movie Database \cite{Pierangeli_2021}.

For image segmentation, prior work has looked at performing real-time image segmentation on an FPGA-based accelerator \cite{ghielmetti2022real}. They demonstrate low latency of about 3ms with less than 30\% resource usage, while maintaining accuracy on the Cityscapes dataset. Prior work has also looked at the use of FPGA accelerator for performing real-time skin segmentation, demonstrating that no more than 88\% of the system resources were used in the process \cite{de2006fpga}. In the context of designing more efficient image segmentation models for accelerators, prior work has developed the Fast Spatial Feature Network (FSFNet) that achieved high throughput and accuracy on the Cityscapes dataset \cite{kim2020accelerator}. Self-driving and defect detection applications rely heavily on their capacity to segment images quickly and accurately. Whilst most research focuses on digital accelerators, we investigate image segmentation on photonic accelerators.

\subsection{Mitigating Quantization \& Analog Noise Error}
Similar to fully digital systems, one of the key challenges of executing DNNs on photonic accelerators stem from the reduced precision supported by the electro-optical components, such as the analog-to-digital converters (ADCs) and digital-to-analog converters (DACs) \cite{basumallik2022adaptive}. While using low precision data types can significantly increase DNN throughput, the model's accuracy degradation can be too large to accomplish the task at hand  \cite{gholami2021survey}. To address this, fine-tuning and post-training quantization techniques have been used to improve the accuracy of models running low precision inference \cite{gholami2021survey}.

Most recently, the approach of dynamically scaling vectors,
such as VS-Quant \cite{dai2021vs} and Adaptive Block Floating Point (ABFP) \cite{basumallik2022adaptive}, have emerged as a potential solution to maintaining accuracy at low precision. Both ABFP and VS-Quant follow a similar approach, by computing scaling factors (i.e., the maximum magnitude) over vectors of length $n$, with VS-Quant adding an additional level of quantization for the scale factors themselves. In this paper, we explore the use of ABFP in photonic accelerators as a method of obtaining good out-of-the-box (OOB) performance for image segmentation. We do not explore the use of a second-level quantization of the scale factors themselves (we leave the scales in \bfloat{}), although the second-level quantization could be utilized in the context of accelerator designs for further improvements \cite{keller202217}. 

Among training methods, for both VS-Quant and ABFP, the authors have explored the use of fine-tuning to further boost model accuracy. For ABFP, the authors propose using Differential Noise Fine-tuning (DNF) as a faster alternative to the typical quantization-aware training approach. In the case of image segmentation models that do not perform well OOB, we explore the use of DNF to recover model accuracy. Note that in this case, fine-tuning is performed off-device and the fine-tuned model is deployed on the device for inference only. 

\subsection{Defect Detection}
Defect detection is used to enhance manufacturing process efficiency, and to ensure the adherence of parts to specifications. Existing methods for defect detection include deep learning for image classification \cite{LI2022429, s19183987}, object detection\cite{AHMAD2022181}, or instance segmentation\cite{cao2020pixel, uzen2023depth, uzen2022swin, lin2023electrode}. Defect detection in real-time on edge devices has particularly gained significant attention for its advantages including reduced latency, improved privacy, enhanced reliability, and efficient bandwidth utilization \cite{NAIN2022588}. Techniques for deploying defect detection on edge devices include approaches like transfer learning \cite{kevin2021federated}, model compression \cite{choudhary2020comprehensive}, pruning and quantization \cite{liang2021pruning} and federated learning \cite{kevin2021federated} to reduce the model complexity and memory usage. However, prior work has not looked at the deployment of defect detection models on photonic devices.

\section{Methods}
In this section, we discuss the implementation details of the photo-core \cite{demirkiran2022electrophotonic} that we use as a reference for our experiments. We then describe the two accuracy recovery methods explored in this work, namely Adaptive Block Floating Point (ABFP) and Differential Noise Fine-tuning (DNF) \cite{basumallik2022adaptive}.

\subsection{Hardware Design}



We investigate the performance of image segmentation DNNs on photonic devices, by evaluating the accuracy, throughput and energy implications of executing the matrix multiplication operations on the photo-core, while performing all the non-linear operations (such as activations) in high-precision digital. Within the photo-core, a weight-stationary approach to computing matrix-vector products (MVPs) is adopted. The weights and input activations of each layer are first unfolded into 2D matrices if necessary (e.g., using kn2row \cite{anderson2017low} for convolutions). Since the photo-core has a fixed size of $n \times n$, weight matrices that are bigger/smaller than $n \times n$ are zero-padded to a multiple of $n$ (called \textit{tile size}) and then tiled into $n \times n$ sized sub-matrices. The input vectors are similarly zero-padded and tiled. Each tile or vector is then loaded onto the photo-core one-by-one. The weight tile is programmed into a Mach-Zehnder Interferometer array, and the input vector is encoded into optical signals. The MVP is then performed via photonics, by modulating the signal strength of the light. The resulting photo-current is converted to digital, via ADCs. The partial results are digitally accumulated for the final output (see Figure \ref{fig:abfp_flow}); please refer to  \cite{demirkiran2022electrophotonic} for more photo-core details.


The chosen DAC precision for inputs and weights is 10 bits and 7 bits, respectively, with an 11-bit ADC at the output. Particularly, we note that the output of the GEMM operations in the photo-core is quantized at the output ADCs, which adds to the error caused by input and weight quantization. This differs from digital devices that retain higher precision in outputs \cite{wu2020integer}. In the following sections, we discuss the implications of this with methods to mitigate its impact in photonic accelerators.


\subsection{Adaptive Block Floating Point (ABFP)}
The inference accuracy of the DNNs on the photo-core is impacted by two key factors: the presence of analog noise (e.g., shot noise, thermal noise) and the quantization noise due to the reduced weight and activation precision at the ADCs. To mitigate this, we explore the use of ABFP \cite{basumallik2022adaptive} for quantizing the weights and activations to the photo-core. ABFP reduces quantization effects by scaling vectors of length $n$
in the weight tiles and input vectors. In addition to per-vector scaling, ABFP also uses an over-amplification factor (i.e., gain $G$) at the output. The use of over-amplification helps increase the signal-to-noise ratio in order to mitigate the impact of analog noise. In our experiments, we analyze this composite framework based on its overall impact in terms of accuracy, throughput and energy.  

Figure \ref{fig:abfp_flow} shows the workflow of ABFP with the photo-core. Given a weight matrix, $N_r \times N_c$, ABFP computes the scale as the maximum values over each row of the $n \times n$ sub-tile, i.e., one scale per row vector of length $n$. The scales are stored in \bfloat{}. Instead of computing the scales over the entire matrix, the use of per-vector scales, $S^W$, for each $n$-sized row helps mitigate the impact of quantization. Similarly, for the input vector, the scales, $S^X$, are computed as the maximum over vectors of length $n$. The weight sub-tiles and input vectors are quantized using the corresponding scales and passed into the photo-core via input and weight DACs. 

\subsection{Simulation Details}
For this work, we designed a simulator for the photo-core, that captures the overall architecture shown in Figure \ref{fig:abfp_flow}, while run on NVIDIA A100 GPUs. Specifically, non-linear operations within the DNNs are computed directly at higher GPU precision, whereas, MVPs that would occur on the photo-core are digitally simulated. The photo-core MVP simulation includes modeling quantization and sources of error that stem from analog noise such as thermal noise and shot noise. These noise sources affect the overall MVP output, and is modeled by adding a cumulative noise term to it. The noise is sampled from a zero-mean normal distribution with a standard deviation that reflects the effects of the cumulative analog noise. Hence, the effects of analog noise and quantization are combined in the digital simulation to model the complete electro-photonic system for DNN inference shown in Figure \ref{fig:abfp_flow}. Within the simulation, the MVP of the weight tile $W$ and input vector $X$ is given by:
\begin{equation}
Y^q = Q(W; S^W, \Delta_W) * Q(X; S^X, \Delta_X)
\end{equation}
The operator, $Q$, is typical quantization operation defined as:
\begin{equation}
Q(X; S^X, \Delta_X) = \nint*{\frac{X}{S^X}\Delta_X} \\
\end{equation}


Where, $\Delta_m = 2^{b_m - 1} -1$ for $b_m$ integer bits, and $\nint{.}$ represents rounding and then clipping between $\left[-\Delta_m, \Delta_m\right]$. We use 7-bit weights and 10-bit inputs, i.e., $b_W=7$ and $b_X=10$. Eqn (2) represents a typical quantization operation, where the inputs are first scaled, then rounded and clipped to the specified range \cite{wu2020integer}. The resultant partial output for a tile (see Eqn (1)) is then amplified via the application of gain $G$ (we use $G=4.0$). To simulate the impact of analog noise, we add a Gaussian noise $\mathcal{E} \in \mathcal{N}(0,\sigma)$ to the resultant output. The noisy result is then quantized at the output ADCs and further de-quantized using the corresponding weight and input scales and converted to \bfloat{} as follows:
\begin{equation}
Y = Q(Y^qG + \mathcal{E}; n\Delta_X\Delta_W, \Delta_Y)\frac{nS^W S^X}{G\Delta_Y}
\end{equation}
Eqn (3) represents 11-bit output quantization ($b_Y=11$), and then de-quantization to \bfloat{}. The final output is obtained by \bfloat{} accumulation over the partial tile outputs. 


\subsection{Differential Noise Fine-tuning (DNF)}
For models that do not perform well OOB, we fine-tune them using Differential Noise Finetuning (DNF) to improve accuracy. In a one-time pre-training step, DNF computes per-layer noise distributions based on the output differences between the quantized and \floatfp{} models. During training, DNF samples noise tensors from the pre-computed noise distributions and adds the noise to the layer outputs of the \floatfp{} model. This way, the noise induced perturbations in the layer outputs enable the models to adapt their weights to quantization error and noise. In contrast to typical Quantization-Aware Training (QAT), DNF uses \floatfp{} precision in the forward pass, eliminating the need for simulated quantization operations in the forward pass thus resulting in significant speed improvements \cite{basumallik2022adaptive}.

\section{Experiments}
\begin{figure}[t]
\centering
\includegraphics[width=0.25\textwidth]{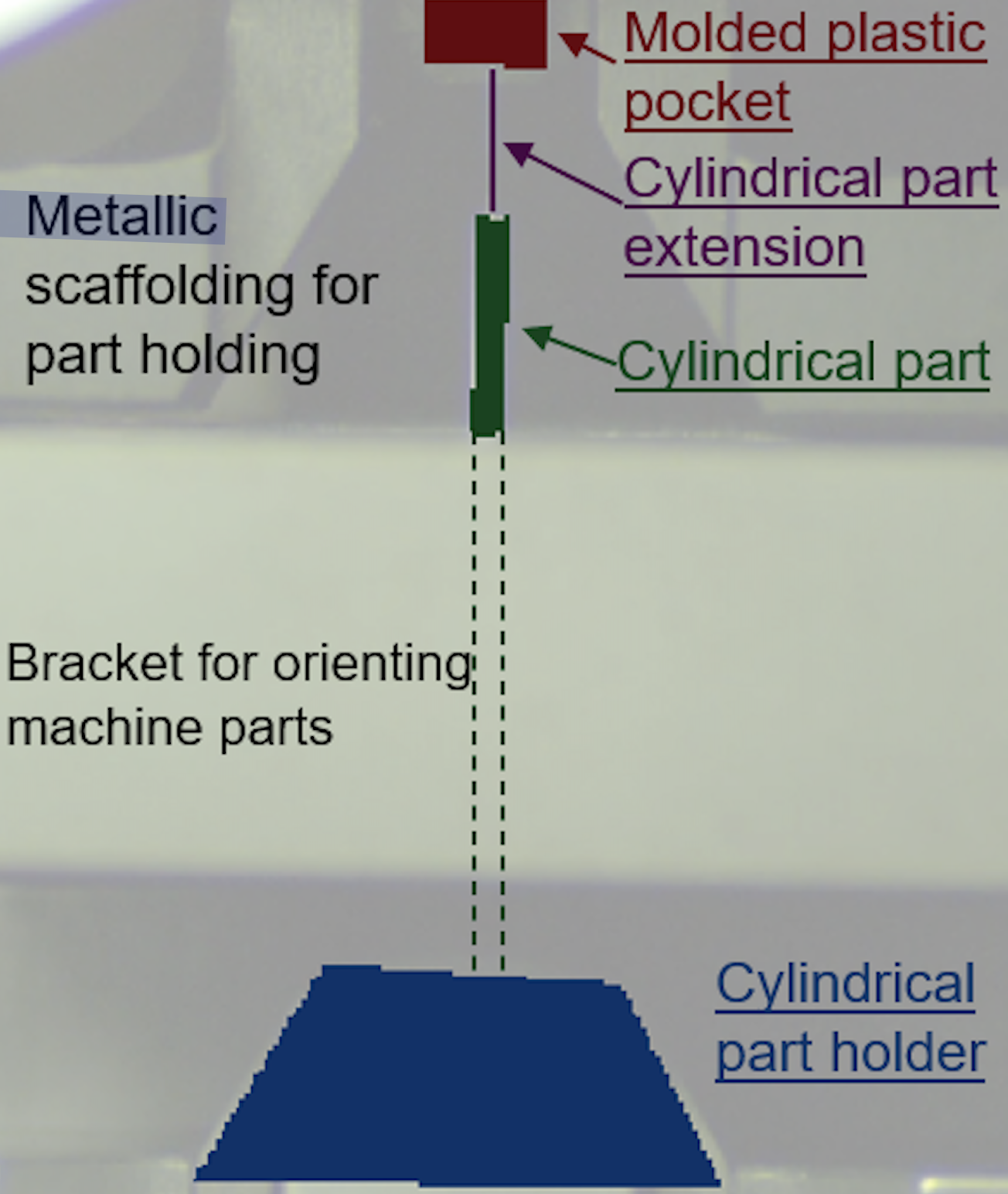}
\caption{An example of the Corning manufacturing process that requires precise insertion of the cylindrical part and its extension to a molded plastic pocket. Semantic segmentation is used to identify four object categories: cylindrical part, cylindrical part extension, molded plastic pocket, and cylindrical part holder. Only the four colored and underlined objects are of interest to the inspection process and are being segmented from the image and other parts. Dotted lines indicate occlusion of the cylindrical part from other components.}

\label{fig:corning_image}
\end{figure}

\begin{table*}[t!]
\centering
\caption{\label{table:results} Experiment results for the models simulated on the photo-core + ABFP (photonic) system with and without DNF.}
\vspace{0.1in}
\begin{tabular}{l l c c c c c c}
 \multicolumn{1}{c}{} & \multicolumn{1}{c}{} & \multicolumn{3}{c|}{\textbf{Pixel Accuracy}} & \multicolumn{3}{c}{\textbf{IoU}}\\ 
  \multicolumn{1}{l|}{Model} &  \multicolumn{1}{l|}{Dataset} &  \multicolumn{1}{c}{FP32} &  \multicolumn{1}{c} {Photonic (OOB)} &  \multicolumn{1}{c|}{Photonic (+ DNF)} & \multicolumn{1}{c}{FP32} & {Photonic (OOB)} &  \multicolumn{1}{c}{Photonic (+ DNF)} \\
  \multicolumn{1}{l|}{} &  \multicolumn{1}{l|}{} &  \multicolumn{1}{c}{} &  \multicolumn{1}{c} {[$\%$ of FP32]} &  \multicolumn{1}{c|}{[$\%$ of FP32]} & \multicolumn{1}{c}{} & {[$\%$ of FP32]} &  \multicolumn{1}{c}{[$\%$ of FP32]} \\
  \hline
\multirow{3}{*}{Mobilenet-dilated-deepsupp} &
    ADE20K &  76.74 & 64.15 [83.6$\%$] & 71.60 [93.3$\%$] & 33.13 & 18.27 [55.2$\%$] & 27.60 [83.3$\%$] \\
    & Cityscapes & 94.45 & 86.99 [92.1$\%$]  & \textbf{94.27 [99.8$\%$]} & 74.64 & 50.89 [68.2$\%$] & 73.69 [98.7$\%$] \\
    & Corning &  97.30 & 67.80 [69.7$\%$]  & 91.42 [93.9$\%$] & 68.76 & 22.00 [32.0$\%$] & 58.28 [84.8$\%$] \\
\hline
\multirow{3}{*}{Resnet50-ppm-dilated-deepsup} &
    ADE20K &  79.09 & 75.49 [95.4$\%$] & 77.46 [97.9$\%$] & 39.82 & 33.14 [83.2$\%$] & 36.21 [90.9$\%$] \\
    & Cityscapes & 93.77 & 74.78 [79.7$\%$] & \textbf{93.29 [99.5$\%$]} & 72.75 & 55.50 [76.3$\%$] & 69.14 [95.0$\%$] \\
    & Corning &  98.03 & 78.07 [79.6$\%$] &  96.79 [98.7$\%$] & 70.51 & 35.05 [49.7$\%$] & 68.27 [96.8$\%$] \\
\hline
\multirow{3}{*}{HRNetv2} &
    ADE20K &  81.45 & \textbf{80.69 [99.1\%]}  & \textbf{80.90 [99.3\%]} & 43.17 & 41.73 [96.7\%] &  42.54 [98.5$\%$] \\
    & Cityscapes & 95.41 & \textbf{95.25 [99.8\%]} & \textbf{95.28 [99.9\%]} & 78.97 & \textbf{78.29 [99.1\%]} & \textbf{78.49 [99.4\%]} \\
    & Corning &  98.49 & 96.89 [98.4\%] & 97.37 [98.9$\%$] & 71.60 & 68.43 [95.6$\%$] &  69.66 [97.3$\%$] \\
\hline
\multirow{3}{*}{Swin-base Maskformer} &
    ADE20K &  82.88 & \textbf{82.15 [99.1$\%$]} & \textbf{82.37 [99.4\%]} & 52.85 & 50.88 [96.3$\%$] & \textbf{52.27 [98.9\%]} \\
    & Cityscapes & 97.42 & \textbf{97.41 [99.9\%]} & NA & 79.18 & \textbf{79.16 [99.9\%]} & NA \\
    & Corning &  98.47 & \textbf{98.39 [99.9\%]} & NA & 71.89 & \textbf{71.85 [99.9\%]} & NA  \\
\hline
\multirow{3}{*}{Swin-large Maskformer} &
    ADE20K &  83.53 & \textbf{83.00 [99.4\%]} & \textbf{82.27 [99.7\%]} & 54.37 & 52.29 [96.2$\%$] & \textbf{53.99 [99.3\%]}  \\
    & Cityscapes & 97.44 & \textbf{97.43 [99.9\%]} & NA & 80.07 & \textbf{79.98 [99.9\%]} & NA \\
    & Corning &  98.52 & \textbf{98.39 [99.9\%]} & NA & 72.23 & \textbf{71.93 [99.6\%]} & NA \\
\hline
\end{tabular}
\end{table*}

We evaluate the performance of image segmentation models for autonomous driving and defect detection on the photo-core. In particular, we evaluate: a) accuracy of different image segmentation DNNs and why certain DNNs perform better than others; b) implications of using the ABFP representation; c) implications of using over-amplification (gain), in terms of accuracy and energy consumption; d) throughput and energy comparisons of image segmentation workloads on the photo-core. We do not make any conclusions about the absolute energy consumption or throughput of the photo-core, relative to other hardware architectures. Instead, our work is focused on empirically analyzing and comparing the DNNs to evaluate their relative strengths on photonic hardware via the photo-core.



\subsection{Datasets}
We evaluate and fine-tune the image segmentation DNNs for autonomous driving using the ADE20k \cite{zhou2018semantic} and Cityscapes \cite{cordts2016cityscapes} datasets, which comprise images of city scenes with 150 and 34 object classes, respectively.

For defect detection, the Corning Defect Detection dataset (not public), consists of RGB imagery from a traditional manufacturing process, involving the insertion of a cylindrical part into a molded plastic pocket (See example in Figure \ref{fig:corning_image}). The objective of the inspection is to verify that positions of the part and the equipment fall within the designated process limits. Pixel-level segmentation facilitates the determination of the relative position of multiple objects and it enables the identification of instances where outcomes deviate from desired specifications. In all three datasets, each pixel in an image is either labeled as one of the classes or as a background pixel.



\subsection{Models}
Five DNNs were chosen based on their sizes and structural nature (convolutional vs. transformers)\footnote{Please see: \url{https://github.com/CSAILVision/semantic-segmentation-pytorch} and \url{https://huggingface.co/facebook/maskformer-swin-large-ade}}. MobileNetv2dilated-c1-deepsup (referred to as mobilenet) is a lightweight CNN architecture designed for efficient semantic segmentation, featuring depth-wise separable convolutions and dilated convolutions to reduce computation costs while maintaining accuracy. ResNet50dilated-ppm-deepsup (referred to as Resnet50) is a variant of ResNet that utilizes dilated convolutions and pyramid pooling to capture contextual information. HRNetv2 is a high-resolution CNN architecture employing multi-resolution fusion to combine features and capture fine-grained information. Maskformer (referred to as swin-base or swin-large) uses a Swin-base or Swin-large backbone with a detection transformer head for instance segmentation tasks, utilizing self-attention mechanisms \cite{cheng2021per}. We used model checkpoints that were pre-trained on the ImageNet dataset, providing a strong initialization for our work.

\subsection{Accuracy Metrics}
We use pixel accuracy and intersection over union (IoU) metrics. Pixel accuracy is the ratio of the number of correctly predicted pixels, over the total number of non-background pixels in the image. The Intersection over Union (IoU) metric measures the overlap between the predicted segmentation mask and the ground truth mask. IoU is calculated as the ratio of the intersection of the predicted and ground truth masks, over their union. The IoU $\in [0,1]$, where 1 indicates a perfect overlap between the predicted and ground truth masks, and 0 indicates no overlap. We measure the average IoU across the object classes in the image and then report the mean IoU (mIoU).  

\section{Results}
\subsection{Accuracy Analysis}
\label{subsec:accuracy}

\begin{table}[t]
\centering
\caption{\label{tab:result_wo_abfp} Comparing input activations with and without using ABFP (vs. using per-tensor max). Using ABFP outperforms full tensor max, particularly in cases where outliers are more prominent.}
\begin{tabular}{c|c|c}
         & \multicolumn{1}{c|}{Without ABFP}                                                 & With ABFP                                                    \\
Model    & \multicolumn{1}{c|}{\begin{tabular}[c]{@{}c@{}}mIoU [\% FP32] \end{tabular}} & \begin{tabular}[c]{@{}c@{}}mIoU [\% FP32]\end{tabular} \\ \hline
Resnet50 & 6.61 {[}9.1\%{]}                                                                  & \textbf{55.5 {[}76.3\%{]}}                                           \\
HRNetv2  & 65.72 {[}83.2\%{]}                                                                & \textbf{78.29 {[}99.1\%{]}}                                          
\end{tabular}
\end{table}

The performance of the image segmentation models in terms of pixel accuracy and mIoU is shown in Table \ref{table:results}. For the Mobilenet and Resnet50 models, we see that the out-of-the-box (OOB) performance is less than 99\% of \floatfp{}. However, HRNetv2 and the Maskformer models (Swin-base and Swin-large) perform well OOB on all datasets, achieving $\simeq99\%$ of \floatfp{}. We further see that in all the cases, the model performances can be improved through the use of differential noise fine-tuning (DNF). DNF enables the Mobilenet and Resnet50 models to achieve significant improvements compared to their OOB performance. Overall, the swin-large maskformer outperforms the other networks in terms of pixel accuracy and mIoU on all the tested datasets.

\textbf{Does ABFP help with OOB model performance?} Table \ref{tab:result_wo_abfp} displays the results comparing the accuracy of HRNetv2 and Resnet50, with and without ABFP. The utilization of ABFP leads to a notable improvement in the final mIoU, particularly for Resnet50 in comparison to HRNetv2.

\begin{figure}[t]
\centering
\includegraphics[width=0.46\textwidth]{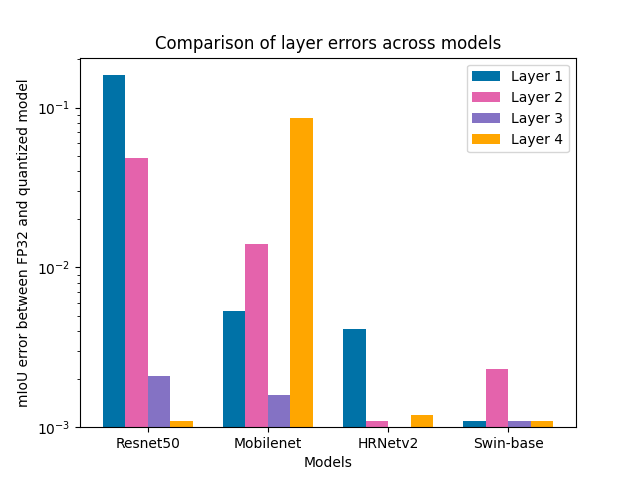}
\caption{Analysis of layer sensitivity. Most of the mIoU drop compared to FP32 mIoU, is accounted for within the first four layers of the models.}
\label{fig:layer_compare}
\end{figure}

\begin{table}[]
\centering
\caption{\label{table:results_outbits} Ablation results for running the most sensitive layer of Resnet50 (layer \#1) on Cityscapes, by toggling different levels of quantization in the photonic system. As shown, output quantization of the sensitive layer results in almost all of the loss compared to FP32.}
\begin{tabular}{c|c|c}
\textbf{Quantization}                & \textbf{Photonic accuracy} & \textbf{Photonic mIoU} \\
    \textbf{Setting}           & \textbf{{[}\% FP32{]}} & \textbf{{[}\% FP32{]}} \\ \hline
With all quantization                         & 74.78 {[}79.75\%{]}               & 55.50 {[}76.29\%{]}                \\
No input quantization           & 74.72 {[}79.68\%{]}               & 55.65 {[}76.49\%{]}                \\
No weight quantization          & 74.22 {[}79.15\%{]}               & 55.42 {[}76.18\%{]}                \\
\textbf{No output quantization} & \textbf{93.45 {[}99.66\%{]}}      & \textbf{71.42 {[}98.17\%{]}}      
\end{tabular}
\end{table}

\begin{figure}[t]
\centering
\includegraphics[width=0.46\textwidth]{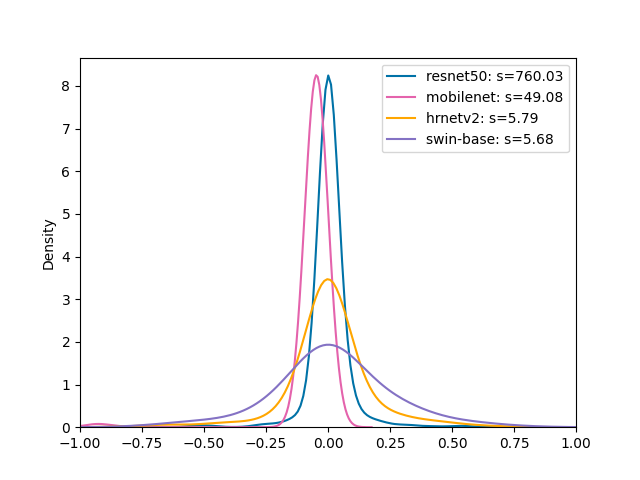}
\caption{Normalized distributions (in range $[-1.0, 1.0]$) of the outputs of the most sensitive layers of the different models. HRNetv2 and Swin-base has a better utilization of the range even for the most sensitive layers, thus resulting in less accuracy drop compared to FP32, on the photo-core. The $s$, in the legend, is the scaling factor from the largest activation.}
\label{fig:layer_hist}
\end{figure}

\textbf{Why do HRNetv2 and Maskformer models have superior OOB results?} In Table \ref{table:results}, we noted that some models under-performed others significantly, in spite of using ABFP. To investigate the reasoning behind this, we performed a layer sensitivity analysis of the models \cite{wu2020integer}. During layer sensitivity analysis, a single layer is quantized at a time, and the model mIoU is evaluated. Layers that result in a lower mIoU when quantized are considered to be ``sensitive'' to loss of precision \cite{wu2020integer}. In Figure \ref{fig:layer_compare}, we show the
results of the layer sensitivity analysis for the first four layers of each model in terms of the drop in mIoU between the \floatfp{} model and the model with the specific layers quantized. We see that most of the drop in mIoU happens within the first few layers of the models. In particular, layer \#1 of Resnet50 and layer \#4 of Mobilenet exhibit increased sensitivity to quantization, i.e., they exhibit the highest mIoU drop compared to the \floatfp{} mIoU. This is consistent with observations on typical digital hardware \cite{joshi2020accurate}. However, for HRNetv2 and Swin-base, even the most sensitive layers only result in a small drop in mIoU. 

To further understand whether the specific layers are sensitive due to input, weight or output quantization, we perform an ablation study by isolating each level of quantization as shown in Table \ref{table:results_outbits}. Specifically, we note that \textit{output quantization} results in almost all of the mIoU loss. The additional quantization at the output ADC introduces a large source of error that impacts the model performance. Following up on this observation, we visualize the normalized output distributions of the most sensitive layers of the different models in Figure \ref{fig:layer_hist}. We note that the maximum output activation value (denoted by $s$) varies across the models, and in particular, Mobilenet and Resnet50 models have very high outliers. A range of $\pm$ three standard deviations of the normalized ResNet distribution utilizes only $23\%$ of the quantization levels whereas the swin-base model uses approximately $69\%$ of the quantization levels. This leads to poor range utilization that in turn results in poor quantization, impacting the performance of Resnet50 and Mobilenet.

\begin{figure}[t]
\centering
\includegraphics[width=0.46\textwidth]{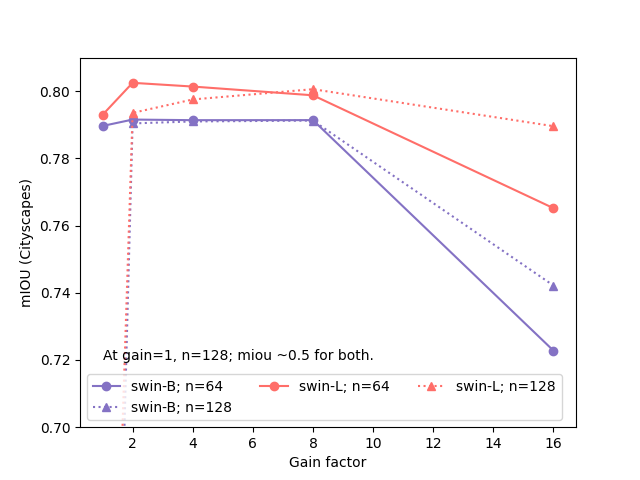}
\caption{Analysis of accuracy vs. tile size $n$ and gain for Swin-base and large Maskformer. The use of gain helps improve accuracy by increasing the signal-to-noise ratio. However, beyond a certain point, gain causes saturation and clipping of activations, resulting in a drop in model accuracy.}
\label{fig:tile_gain}
\end{figure}

\textbf{How does gain help with OOB performance?} To better understand the impact of the chosen tile size, $n$, and gain, particularly for the models that achieve 99\% accuracy OOB (Swin-base and Swin-large Maskformer), we investigate this relationship in Figure \ref{fig:tile_gain}. We make the following observations: a) as $n$ reduces, mIoU generally improves, since smaller tile sizes can further mitigate the impact of outliers which is a key issue as previously described; b) the use of gain helps improve mIoU until a certain point, since the signal-to-noise ratio is correspondingly improved; c) beyond a threshold, gain causes the mIoU to drop due to saturation. Saturation of values (when amplified by the gain) causes clipping of critical activations thus impacting model accuracy. Overall, the combination of ABFP + gain with specific choices of $n$, enable models like Swin-base and large Maskformers to achieve good OOB results.

\subsection{Energy Usage Analysis}


\begin{figure}[t]
\centering
\includegraphics[width=0.46\textwidth]{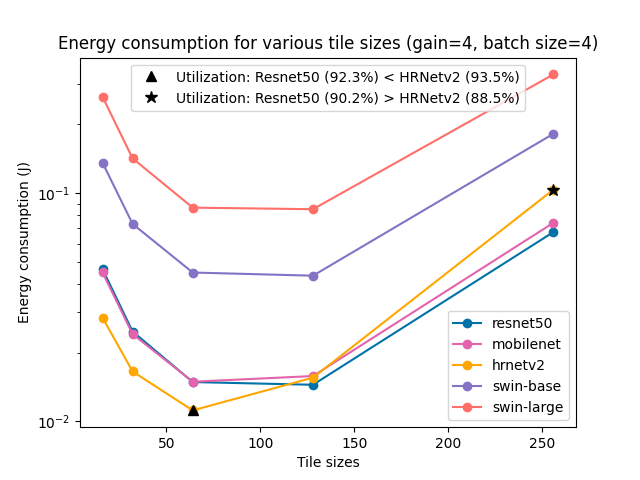}
\caption{Analysis of the energy consumption of one batch of data for different tile sizes. Hrnetv2 has lesser consumption than Resnet50 at $n=64$ (denoted by $\blacktriangle$) compared to $n=256$ (denoted by $\star$) due to a decrease in photocore utilization relative to Resnet50.}
\label{fig:energy_graph}
\end{figure}

The laser power of the photo-core depends on the gain $G$, tile size $n$, and fixed internal parameters such as MZI loss, laser efficiency and coupling loss. The key factors of interest in this work are $n$ and $G$, and we analyze the \textit{relative} scaling behavior of photo-core energy consumption for the image segmentation workloads based on the two parameters. This also allows us to abstract the remaining parameters from our analysis since they are fixed across the models (MZI loss, laser efficiency etc.). With this form of abstraction, we can group the internal laser parameters to two sets of constants, $\alpha$ and $\beta$. Then the photo-core power $P$, can be summarized in terms of $G$ and $n$:
\begin{equation}
P(G,n) = \left(G\alpha^n + \beta\right)n
\end{equation}
The total amount of photo-core energy required for a given DNN inference at a specified gain and tile size is then $E(G,n) = T*P(G, n)$, where $T$ is the total time taken to execute the DNN workload (\# of MVPs $\times$ time taken for the photo-core per MVP), plus data transfer costs for the weights to be sent and loaded onto the photo-core (set to 40ns, 10ns respectively)\footnote{Since we focus primarily on the photocore, we exclude energy analysis of digital aspects such as quantization, and data transfer costs of inputs.}. This abstraction captures the exponential relation between energy and $n$, and the linear relation between energy and $G$ (with a slope that is determined by $T$ and $n$). Empirically, for our system, a gain of 2 increases energy consumption by $1.4\times$ for $n=64$ and $1.9\times$ for $n=128$, compared to using no overamp, indicating that the accuracy improvements from Section \ref{subsec:accuracy} (Figure \ref{fig:tile_gain}) come at an energy cost. 


Figure \ref{fig:energy_graph} shows how the relative energy consumption scales as a function of tile sizes for a single batch of data. We keep the other photo-core parameters fixed, and do not focus on the absolute energy consumption of the models. Instead, we make the following observations: a) the swin-base and swin-large maskformers consume $\simeq2\times$ and $\simeq4\times$ more energy respectively, than CNNs; b) the energy consumption reduces up to $n=64, 128$. This is because smaller tile sizes ($<64$) require more MVPs and correspondingly larger run-times. Beyond $n=128$, the $\alpha$ parameter begins to dominate for larger tile sizes leading to more energy consumption; c) the energy consumption of HRNetv2 is less than Resnet50 at $n=64$ (shown as $\blacktriangle$), but higher at $n=256$ (shown as $\star$). This is due to the reduced photo-core utilization of HRNetv2 compared to Resnet50, at $n=256$. Depending on the matrix sizes in these models (that are not divisible by tile size), larger tile sizes will lead to under-utilized tiles and higher energy consumption. 
  


\subsection{Throughput Analysis}
\begin{figure}[t]
\centering
\includegraphics[width=0.46\textwidth]{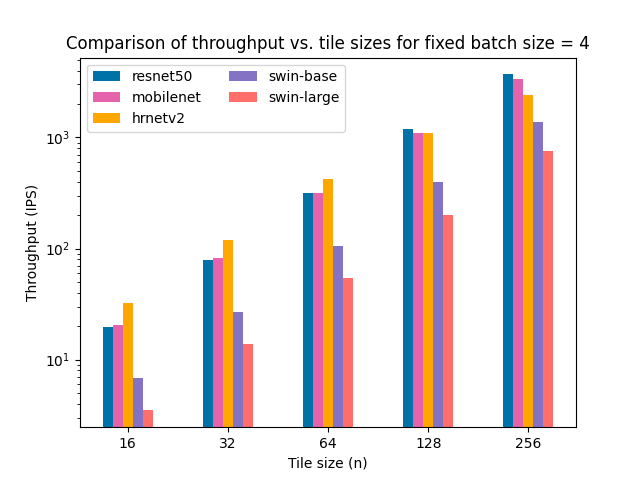}
\caption{Analysis of throughput vs. tile size $n$. Smaller tile sizes can mitigate impact of outliers, but reduce throughput significantly compared to larger tiles.}
\label{fig:throughput}
\end{figure}

Figure \ref{fig:throughput} shows the throughput for the different DNNs for different tile sizes and a fixed batch size of 4. We see that: a) swin-base and swin-large maskformers have lower throughput compared to the CNNs. This highlights a trade-off with the maskformer models that have better OOB accuracy results, but lower throughput compared to CNNs; b) although smaller tile sizes offer accuracy improvements as shown in Section \ref{subsec:accuracy} (Figure \ref{fig:tile_gain}), it results in lower throughput. Hence, there is an accuracy-vs-throughput trade-off when selecting the appropriate tile size to use for the image segmentation workloads.

\section{Discussion}
In this work, we presented an analysis of image segmentation workloads for autonomous driving and defect detection on photonic hardware. We analyzed different convolutional and transformer models, and a summary of our findings is captured in Figure \ref{fig:radar}. We see that while the maskformer models achieve good OOB accuracy, they are less competitive in terms of throughput and energy consumption. In contrast, HRNetv2 (particularly \textit{with} DNF) presents a good trade-off in terms of energy, throughput and accuracy. Thus, \textit{HRNetv2 is a promising candidate for image segmentation on photonic devices}. We highlight some additional key insights from this work:
\begin{enumerate}
    \item The accuracy of image segmentation on photonic hardware is significantly impacted by the output quantization. While both ABFP and DNF are promising for improving accuracy, techniques such as parameterized activations to mitigate outliers \cite{choi2018pact}, can also be helpful.
    \item Accuracy degradation mainly occurs due to relatively few layers of each model. Existing approaches that execute sensitive layers at higher precision while keeping others lower \cite{wu2020integer, khoram2018adaptive}, can be beneficial here. Photonic systems should consider supporting this hybrid approach with controllable bit-precisions for sensitive layers.
    \item There is a complex relationship between the photo-core tile size, gain, and DNN architectures, influencing the energy and throughput profiles. Photonic systems may be designed to dynamically adjust internal parameters such as gain to optimize energy efficiency. DNN architectures can also be re-designed as in \cite{zoph2016neural}, to be optimal for the specific photo-core tile size to improve throughput.
\end{enumerate}

\textbf{Drawbacks and future work:} While we have looked at hardware parameters like tile size and gain, we have not analyzed other architectural aspects of the hardware in-depth, such as wall-plug efficiency of the laser, data storage or movement, and non-linearities on digital ASICs. We have also not explored other potential throughput optimizations of attention layers \cite{shoeybi2019megatron} that may improve Maskformer performance over CNNs. We explore our work in the context of the photo-core; however, we believe that our findings with ABFP, DNF, tile sizes and gain will offer insights for future work in algorithmic and hardware enhancements to photonic accelerators for ML workloads.

\section{Acknowledgements}
The authors would like to thank colleagues Cansu Demirkiran and Alexander Sludds for their helpful insights and discussions. We also thank our reviewers for their constructive feedback.



\bibliography{references}
\bibliographystyle{unsrt}

\end{document}